# A Distributed AI Aided 3D Domino Game.

Şahin Emrah Amrahov , Orhan A. Nooraden

Department of Computer Engineering, Engineering Faculty of Ankara University, Aziz Kansu Building, Tandogan Kampus, Ankara, 06100 Turkey (e-mail: emrah@eng.ankara.edu.tr).

## Abstract

In the article a turn-based game played on four computers connected via network is investigated. There are three computers with natural intelligence and one with artificial intelligence. Game table is seen by each player's own view point in all players' monitors. Domino pieces are three dimensional. For distributed systems TCP/IP protocol is used. In order to get 3D image, Microsoft XNA technology is applied. Domino 101 game is nondeterministic game that is result of the game depends on the initial random distribution of the pieces. Number of the distributions is equal to the multiplication of following combinations: $\binom{28}{7}\binom{21}{7}\binom{14}{7}$. Moreover, in this game that is played by four people, players are divided into 2 pairs. Accordingly, we cannot predict how the player uses the dominoes that is according to the dominoes of his/her partner or according to his/her own dominoes. The fact that the natural intelligence can be a player in any level affects the outcome. These reasons make it difficult to develop an AI. In the article four levels of AI are developed. The AI in the first level is equivalent to the intelligence of a child who knows the rules of the game and recognizes the numbers. The AI in this level plays if it has any domino, suitable to play or says pass. In most of the games which can be played on the internet, the AI does the same. But the AI in the last level is a master player, and it can develop itself according to its competitors' levels.

## Introduction.

Starting from C.Shannon's paper[1], game artificial intelligence investigators have developed computer games that can play against to human. Deep Fritz, Chinook, Othello, Scrabble game programs can be exemplified. Although all of these programs are strong, none of them are perfect and they cannot win unless the competitors make a mistake. Today game programs make progress fast in terms of graphic and visual quality. However, it is early to say the same thing for artificial game intelligence. Actually game artificial intelligence concept is not defined clearly. Game program developers and academicians understand different things from this concept. Academicians think about behaviors of the characters in the game when they consider the game artificial intelligence. On the other hand, for game developers game AI is used to encompass techniques broadly. Path finding, animation systems, level geometry can be exemplified to these techniques. There are significant differences between natural intelligence (human brain) and artificial intelligence[6-9]. Firstly, human can play different games reasonable by using his/her brain. A computer can play different games reasonable only if artificial intelligence for each game is produced. There may be some common properties of games, Allis [2] categorized games according to their common properties. But, these common properties do not save us from writing different computer codes for each game. Secondly, we cannot expect a human to play perfectly, human can make mistakes. In contrast, all computers are expected to play perfect. Actually for a computer to play perfect, game has to be solvable, has to contain no uncertainty and has to have an algorithm which is not NP-hard. For example, checkers is a solvable game[3,12]. In this paper we examine the domino game. With domino various games can be played and various mathematic problems can be solved [4-5,13,15,16]. The game we examine is "101" game. This game contains uncertainty and it is played with 4 players [10-11]. We examine the game in terms of three points. 1) Artificial intelligence 2) Networking environment 3) Visual

Classic domino set consists of 28 dominoes (pieces). The height of each piece in the form of a rectangular prism is less than its other dimensions, and the front face of each domino is divided into two squares by a line. In each square the numbers from 1 to 6 are shown or empty. Back side of the dominoes is designed in the way that they cannot be distinguished from each other.

Various games can be played with domino set. One of the most popular kinds of these games is 101 game. 101 game is played by two people or four people. In this article the four-person game is considered. The rules of the game are as follows:

Around a rectangular table four players as A, B, C, D sits. A and C and B and D are partners. All pieces are mixed and distributed among the players as seven pieces to each player.

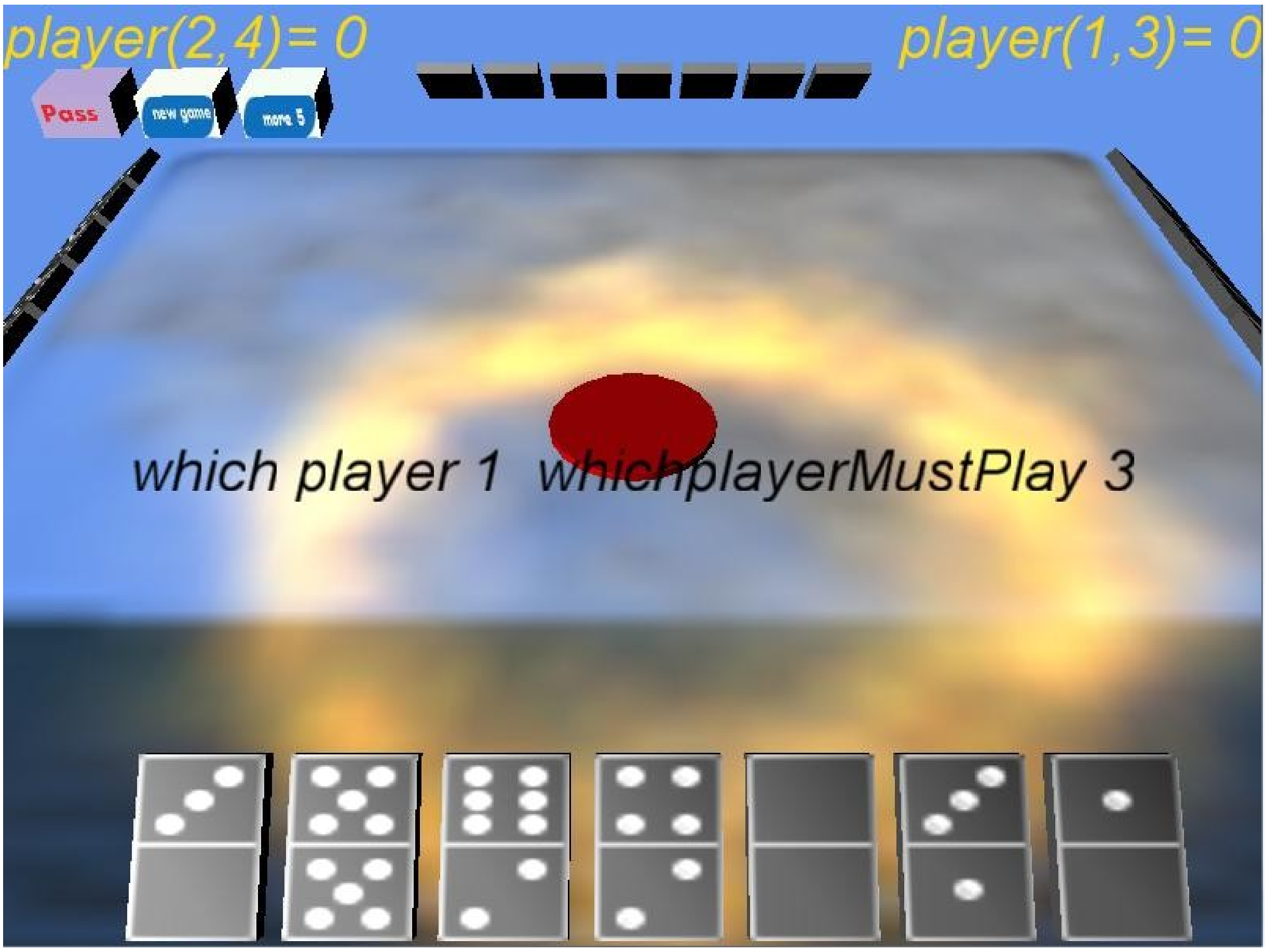

If a player has five or more double stones in his/her hand, that is two sides of a piece are the same,

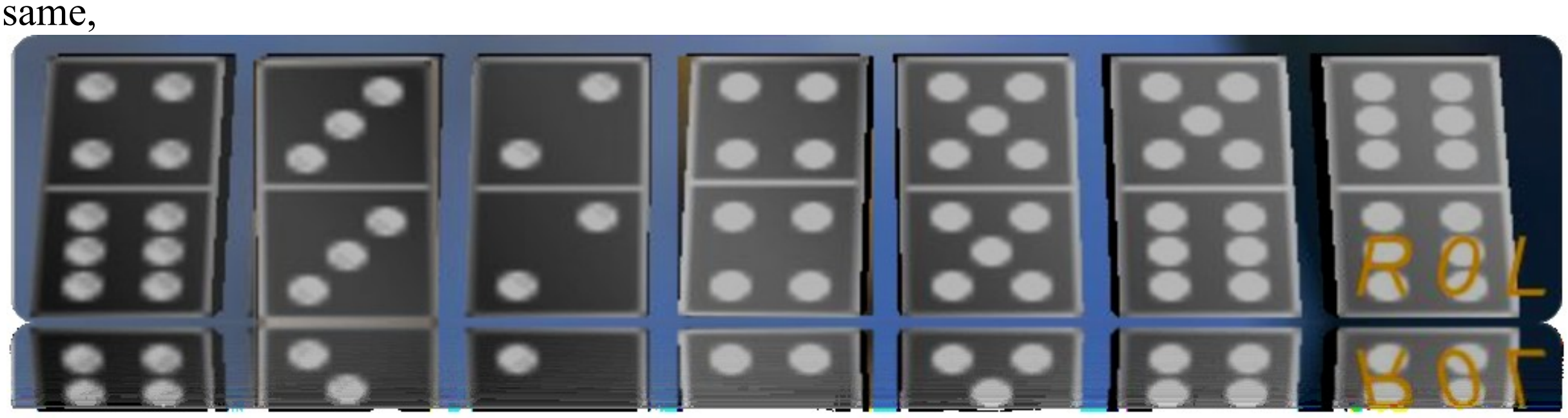

If one player has 6 or 7 dominoes a face of which is the same, then the player informs the situation and shows the stones, and then pieces are mixed again and re-apportioned.

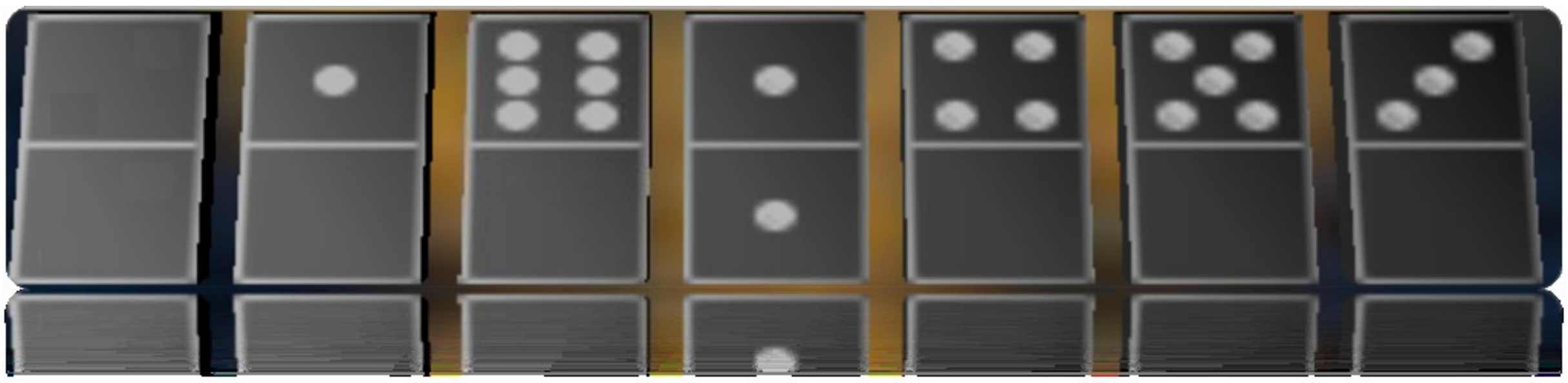

If a player has 5 dominoes a face of which is the same, not having their doubles, the dominoes are mixed again and re-apportioned. After the first distribution of the dominoes if a player violates the rules of the game, he/she and his/her partner are considered lost. In the first game the player who has 1-1 dominoes in his/her hand starts the game, and then the players spurt one by one in the way of forming a chain with the dominoes in the clockwise. If the player has no appropriate domino to play, he/she says pass. If a player says pass, although he/she has a suitable domino to play, the player and his/her partner is considered lost as soon as the situation is revealed. One of the partners who won the previous game starts the next game after the first game after a short-time counseling without disclosing his/her dominoes to his/her partner . When one of the players finishes all the pieces, both he/she and his/her partner win and the game is over. The dominoes of the defeated team are opened and then the numbers on these dominoes are counted. This total is written on the winner's board as achievement. After the games the team who has gathered 101 or more points wins the game. If no one has suitable domino during the game, the game is called as "closed". In a closed game all the pieces of the players are opened, and the numbers on the dominoes are counted. The partners who have less than the other team win as much as the total that the other team has got. The player who threw the last stone which is not double or his/her partner starts the next game.

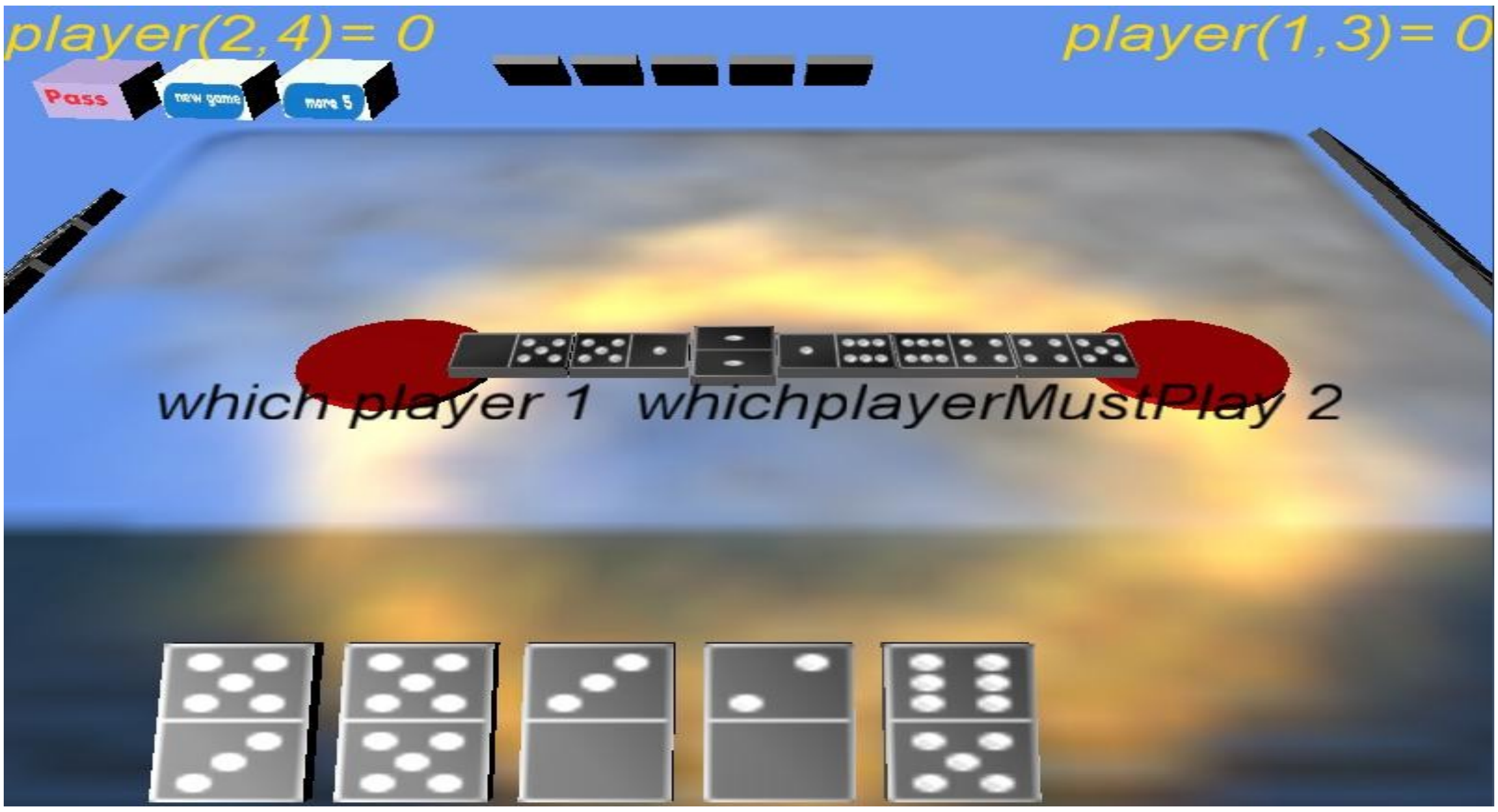

## Network and TCP / IP

Two people can decide which language they will use and can communicate with each other. They can choose English, Turkish or Arabic, but both have to use the same language. Computers work in the same way. TCP / IP (Transmission Control Protocol / Internet Protocol) represent a language that can be used by computers.
TCP / IP are a set of rules and it provides communication and data transfer of two computers with each other.
This set of rules is called the protocol. There are a great number of protocols tasks of which are different. By grouping a number of protocols, a work can be done. The protocols working together are called Protocol Stack or Protocol Suite.
TCP / IP is a widely-used network protocol. Some reasons of its wide usage are its being used in internet, the biggest network of the world, its being compatible with almost all computers of the world and its being supported by main operating systems. TCP / IP is divided into four layers and each layer has a specific task.

## TCP / IP Layers

### Application Layer:

It is at the first level of TCP / IP Layers. It contains all the applications and utilities and provides entry into the network. The protocols at this level enable the user to prepare his/her private information and to transfer data. For example, one of these applications is the HTTP (Hypertext Transfer Protocol) and its task is to transfer the files which create sites and web pages. As a second example, we can give the FTP (File Transfer Protocol). With the help of this protocol, the files are transferred over the network.

### Transport Layer:

In this layer the opportunity of connection and guarantee of communication are provided. We can give the TCP (Transmission Control Protocol) as an example to these protocols. TCP makes us sure that the data is transmitted. TCP is of Connection-based type. Namely, it needs to establish a session before transferring the data between computers. Moreover, it provides sequential and correct data transmission. If the data is not transmitted, it tries to resend or sends other data package.
The second protocol, UDP (User Datagram Protocol), is of Nonconnection-Based type. It does not establish a session before data transmission between the computers. All of the data which are sent in this type cannot be guaranteed to reach.

### Internet Layer:

This layer has such functions as IP addressing and choosing the best way. It contains the basic protocols. The first is the internet protocol (Internet protocol-IP). This protocol is considered as the most important protocol, because addressing element is found in it. Thus, it gives a specific, unparalleled number to each computer on the network. This number is called IP Address. IP feature directs (routing) and then combines and decompose the package (Packaging). The second basic protocol is (ARP-Address Resolution Protocol). The IP specifies the location of the protocol in the network and by using MAC address it finds destination; ARP protocol searches the address in its memory. When it finds, it shows the way to the address clearly. The third protocol is Internet Control Message Protocol-ICMP. The fourth is Internet Group management protocol –IGMP.

**Network Internet Layer**

It puts the data, requested to send to Network Medium and it retrieves data from the counter side. It contains all physical structures. It also includes the protocols which indicate the way of sending data. For example, A.T.M, Ethernet, Token Ring.
Port IP:
IP is a special number that identifies the program on the network. It is defined in TCP or UDP. Its value is between 0-65535. Some numbers are already set for some programs, we cannot use these numbers. For example, FTP uses port 20th or 21st port and HTTP uses port №80.

**How It works:**

In order to send the message in the application layer on the network, it is transformed into the ASCII Code, then to Binary Code or bite from the text version, then it is prepared by being written on the matrix and is sent with socket. Socket connects Transport, Network and Internet layers. Destination in Network Layer ties up Port Number in Transport with IP. Server Client on the other side approves the invitation. Socket in the server only connects Port Number with Socket Option; then it starts to listen to the port.

**Threading**

The concept of multithreading supports the concept of multitasking of operating system. In this way, any program can run in one thread separately from other threads. In addition to it, the same program can contain more than one thread.
Each program takes a certain amount of place in the processor with regard to its requirements. Operating system takes into consideration the requirements and priorities of the running programs and divides the processor

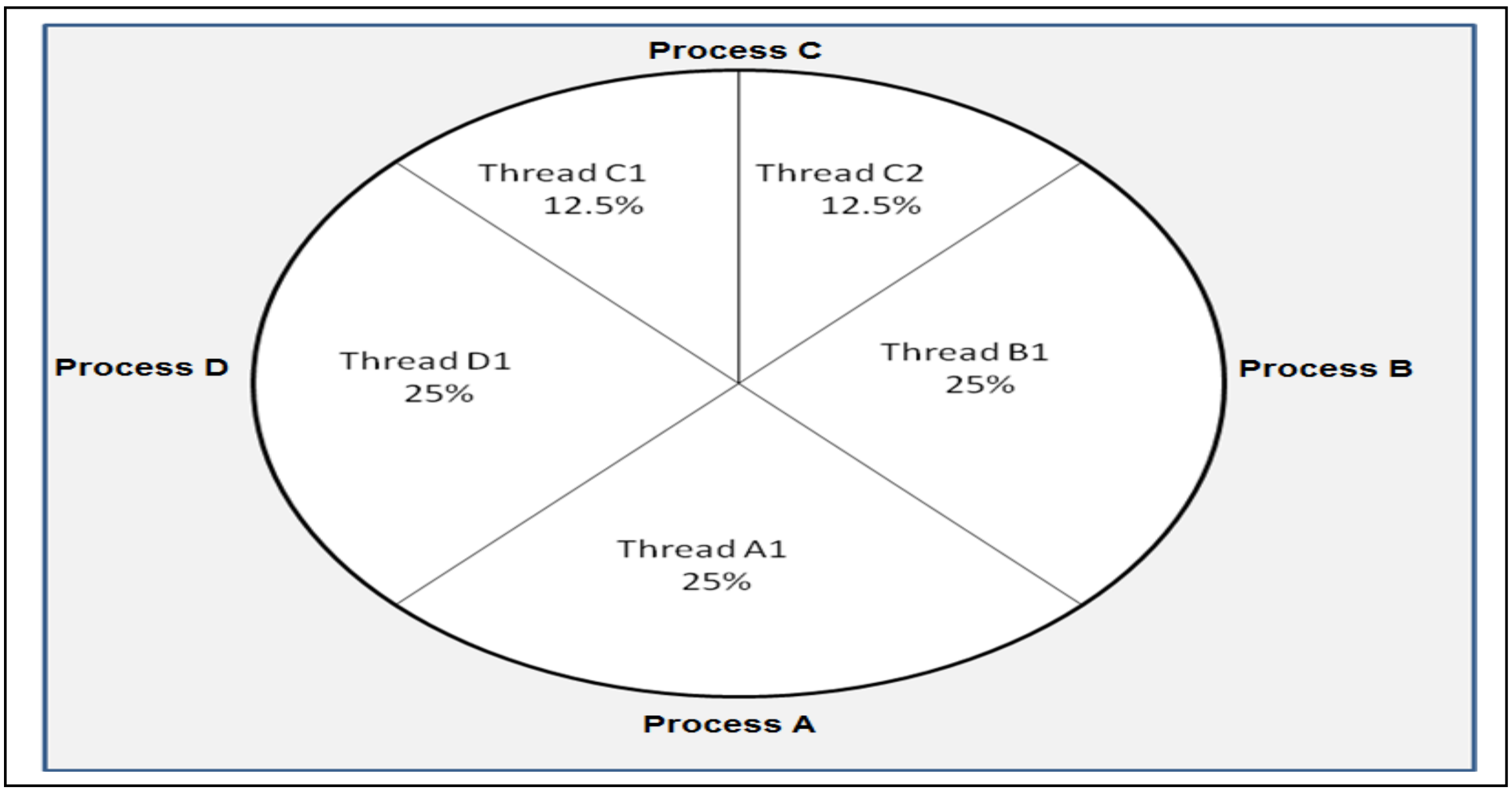

Moreover, we can control the threads in the program. For example, we may suspend it for a while or cancel it by aborting. All these processes can be used freely by the programmer.

### Threads in the software languages

In order to explore the information which is sent in the network programming, it consistently controls the specified port and it runs in an infinite loop. Therefore, it is not possible for us to make any process (to transact) before the effective loop is not ended. Since we use an infinite loop, the transaction will never exist. Solution of the problem is putting the function, which controls the port, into another thread. This is called multithreading. Operating system divides/allocates the processor into two. One of the parts is for the main program, and the other one is for the endless loop. Thus, the functions within the program will not run in stages/steps, but in parallel.

### How we used thread:

4 computers are connected to the server. The server gives a number to the connected network; this number identified the player. The thread is created for every connected network. This thread waits for information from the port, and sends it to other computers on the network. Computer transmits every player's behaviors to other computers. For example, it gathers the player's domino information (location, number, which the player is playing) and sends it to other players' computers on the network by packaging the game information. Other players' computers receive the package. And then, what the player is playing is now played in these computers and it is displayed on the monitor.

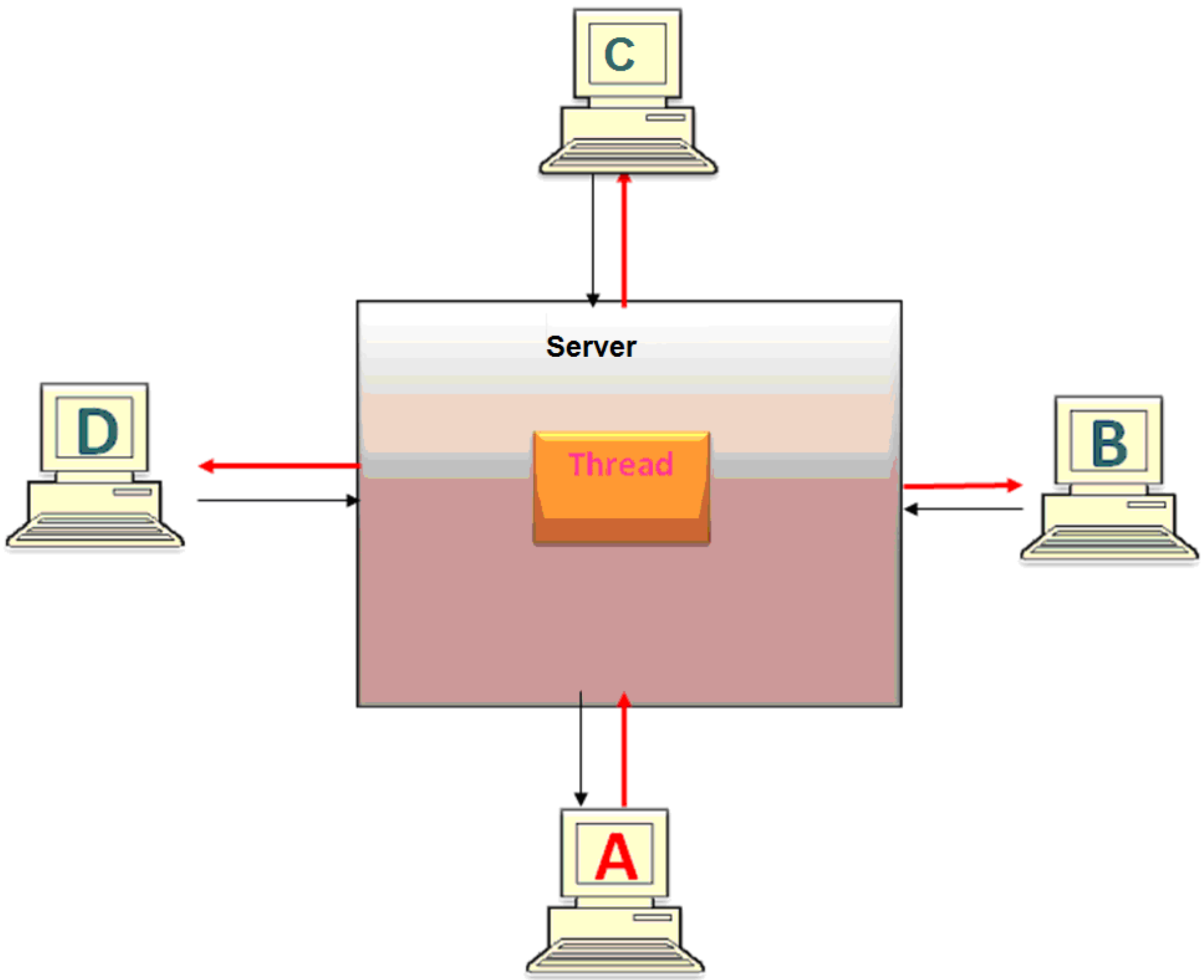

## 3D Domino pieces and the game table

### XNA Description

XNA is a library developed by Microsoft in order to develop games. Visual Studio works at C #. It is more developed/advanced version of DirectX. This library contains many libraries such as graphics, sound library, etc. Games that are made in this library work in XBOX and PC. Input devices (keyboard, mouse, gamepad, game controller) are easily transmitted and the displayed card is reached through XNA. We can also control volume in XNA

## Camera

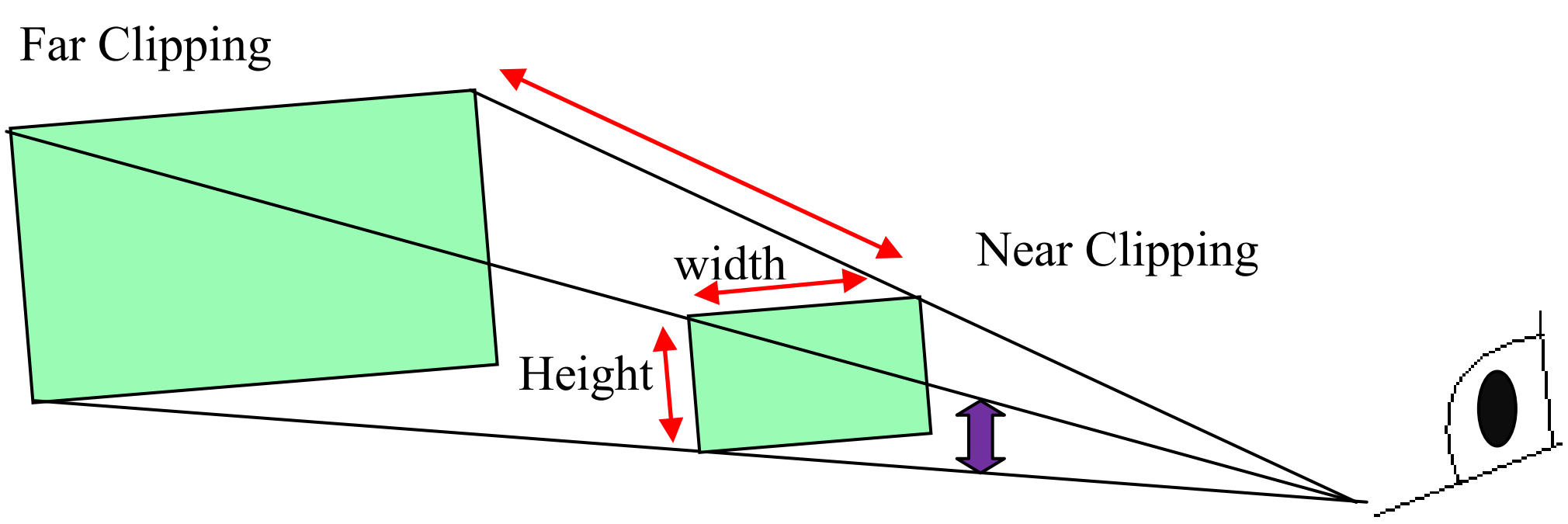

### Domino Design

Domino piece is a cube with two large faces/sides. One of them is the visible face and shows the domino numbers. The other side is not visible. Cube is formed from the triangles in XNA. The cube has eight corners and six faces. Each side consists of two triangles.

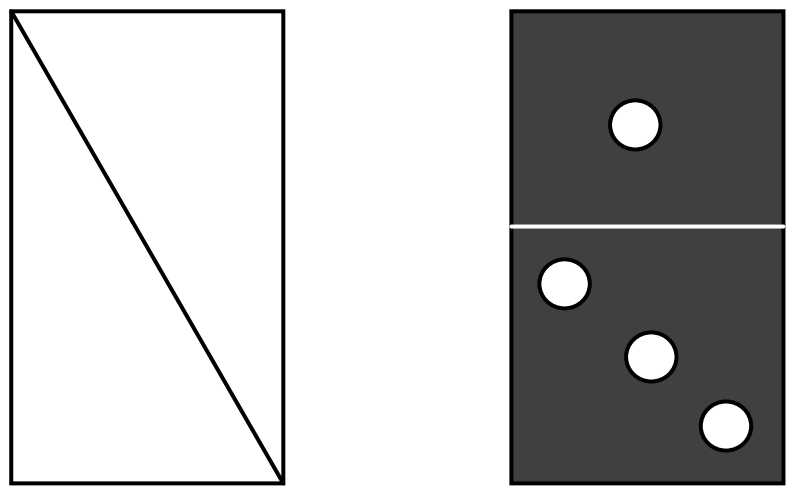

## Artificial intelligence

AI keeps what the players, including itself, are playing until that move in its mind. It looks at the table and at its hand, and if it has one domino to play, it plays without thinking. If it has got some dominoes more than one to play, it decides what domino it will play by thinking. It tries to estimate the dominoes which are left in the players' hands according to the dominoes which the players have played so far. Game strategy is as follows.
1. It tries to open the number which is left more in its and its partner's hands,
2. It tries to throw the domino, whose addition is bigger,
3. When the numbers in its and its partner's hand contradict, it tries to play according to anyone, either it or its partner, who has lesser dominoes in number,
4. It tries to open the number which is not available in the following player's hand, and this time, it wants its partner to have this number in his/her hand,
5. It follows the double dominoes, and it tries to play the double domino which is opened more than three times,
6. If it has the last of the number which is opened on the table in its hand, it does not play this domino unless it is forced to play it,
7. If it has all of the last numbers of one domino of the opened dominoes in its hand, both teams try to do/reach this number,
8. If there is the possibility of closing, it tries to collect the numbers left in its hand and the numbers on the dominoes which it estimates to be left in its partner's hand. And also it tries to collect the numbers which it estimates to be left in its competitors' hands. If it decides that it and its partner have less, it tries to close the game.
9. It tries to play on a number which its partner opens.

### Estimation phase

1. When the player says pass, AI understands that the numbers which are in the ends of the chain on the table are not available,
2. If a player plays insistently plays on the domino which he/she has opened, it estimates that this player does not have the domino which is in the other end of the chain,
3. If a player plays on the number which his/her partner has opened, it estimates that this player does not have the other domino which is in the other of the chain.

### 1st AI-level

In this level, if AI has a domino to play, it chooses and plays randomly.

### 2nd AI-level

Let's take the players as A, B, C, D as players and D as the computer. Let A be the next player from the computer.
The algorithm of D is as follows: It tries to open the domino which is not found in A's hand, and to close the number which is bigger than its in A's hand.
It is doing this in this level as follows:
At the beginning of the game it distributes the dominoes to the players' hands randomly (without violation of the rules), and according to this distribution it begins to play. As the moves go forward, it puts the dominoes into the true players' hands, and takes a random domino from this player and puts the domino which has been played into the hand of the player who has just played.

### The functions used in the 2nd level

Double-domino control: If it is the AI's movement turn, it counts how many dominoes have been opened in the number of the double dominoes. If more than 3 double dominoes have been opened, it marks the double domino as **.
Maxmin function: If the movement turn is on AI, it scans all the possibilities which it can play, and for every possible movement, it subtracts the number on which it estimates to be on the dominoes in A's hand from the numbers of the dominoes in its hand among the numbers will be left as open-ended on the table. According to the differences which it finds it aligns the movements from the biggest to the smallest. When equal differences occur, it chooses one of them randomly and puts it to the top.

### The algorithm of AI at the 2nd level

1. Compulsory Movement 1: It says pass if it does not have a suitable domino to play,
2. Compulsory Move 2: If there is only one domino, it plays,
3. There is a double domino including ** : If they are more than one, it plays the biggest one,
4. If there is no domino including **, it plays the domino on the top as a result of Maxmin function.

### The 3rd AI-level

In addition to the functions at the second level, let's add sum 7 control function at this level.
Sum 7 control: It looks at its hand and at the table. Let's take the ends on the table as A and B. It counts the dominoes which is opened on the table and which it has the dominoes including A. As a result of this counting, if it gets the sum 7, it marks all the dominoes including A as *. It does the same things for B.

### 3rd AI-level algorithm is as follows:

1. Compulsory Movement 1: It says pass if it does not have a suitable domino to play,
2. Compulsory Move 2: If there is only one domino which it can play, it plays it,
3. There is a double domino including ** : If they are more than one, it plays the biggest one,
4. If it does not have any double domino including **, it plays according to Maxmin but the domino to be played as a result of Maxmin has *, it passes to the second domino, and if this domino also has *, it passes to the next domino. If there is no domino including *, it plays the first domino in the Maxmin turn.

### 4th AI-level

At this level, let's replace Maxmin function with Mmaxmin function which is a bit more developed.
MMaxmin function: If the movement turn is on AI, it scans all the possibilities which it can play, and for every possible movement, it adds the numbers which it estimates in its partner's hand to the number which is in its hand among the numbers which will be left as open-ended. It subtracts the number which it estimates on the dominoes in A's hand from the number this multiplication. When equal differences are obtained, it puts the domino which will not be played to its partner's domino on the top. If its competitors have opened both of the ends on the table, it puts the domino which the player's domino on the domino which will be played on the top.

### 4th AI-level algorithm

1. Compulsory Movement 1: It says pass if it does not have a suitable domino to play,
2. Compulsory Move 2: If there is only one domino which it can play, it plays it,
3. There is a double domino including ** : If they are more than one, it plays the biggest one,
4. If it does not have any double domino including **, it plays according to Mmaxmin but the domino to be played as a result of Mmaxmin has *, it passes to the second domino, and if this domino also has *, it passes to the next domino. If there is no domino including *, it plays the first domino in the Mmaxmin turn.

## References

1. C. Shannon, Programming a computer for playing chess, Philosophical Magazine 41(1950), pp.256-275

2. V. Allis, Searching for Solutions in Games and Artificial Intelligence, PhD thesis, Department of Computer Science, University of Limburg, 1994

3._J. Schaeffer, N. Burch, Y. Björnsson, A. Kishimoto, M. Müller, R. Lake, P. Lu, S.Sutphen, Checkers Is Solved, Science, 317(2007), pp.1518 -1522.

4. M.Gardner ,Dominono, Computers and Mathematics with applications ,39(2000), pp.55-56

5. Y.Etzion-Petruschka, D. Harel and D.Myers, On the solvability of domino snake problem,Theoretical Computer Science 131 (1994), pp 243-269

6. A. Wolff, P. Mulhollanda, Z.Zdrahal, R. Joiner, Re-using digital narrative content in interactive games, Int. J. Human-Computer Studies 65 (2007), pp. 244–272

7. R. Metoyer , S. Stumpf , C. Neumann , J.Dodge , J. Cao , A. Schnabel, Explaining how to play real-time strategy games, Knowledge-Based Systems 23 (2010), pp. 295–301

8. M.Ponsen, P. Spronck, H. Munoz-Avila, D. W. Aha, Knowledge acquisition for adaptive game AI, Science of Computer Programming 67 (2007), pp. 59–75

9. N. Halman, Simple stochastic games, parity games, mean payoff games and discounted payoff games are all LP-type problems, Algorithmica ,49(2007), pp.37-50.

10. M.H. Smith, A learning program which plays partnership dominoes, Communications of the ACM,16 (1973), pp. 462 - 467

11.Y.A. Pervin, Algorithmization and programming of the game of dominoes. In Problems of Cybernetics 111, Pergamon Press, New York, 1962, pp. 957-972.

12. A.L. Samuel, Some studies in machine learning using the game of checkers, IBM Journal of Research and Development, 3 (1950), pp. 211-229.

13. S.Reisman, Dominoes: A computer simulation of cognitive processes, Simulation attd Games 3(1972), pp.155-163.

14.A.L. Samuel, Some studies in machine learning using the game of checkers: I1. Recent progress, IBM Journal of Research and Developments, 11 ( 1967), pp.601-618.

15. B.S. Chlebus, Domino-tiling games, Journal Computer System Science, 32(1986), pp.374-392.

16. E.Gradel , Domino games and complexity, SIAM Journal on Computing.19(1990), pp.787-804